\documentclass[letterpaper, 10 pt, conference]{ieeeconf}  

\IEEEoverridecommandlockouts                              

\usepackage{amsmath} 
\usepackage{amssymb}  
\usepackage{amsfonts}
\usepackage{mathtools}
\usepackage[noadjust]{cite}
\usepackage{booktabs,multirow}
\usepackage{gensymb}
\usepackage{array}
\usepackage{dblfloatfix}
\usepackage{makecell}
\usepackage{xcolor}
\usepackage[caption=false]{subfig}
\usepackage{diagbox}
\usepackage{tikz}
\usepackage{longtable}
\usepackage{adjustbox}
\usepackage{booktabs}
\usepackage{xltabular}
\usepackage{multirow}
\usepackage{supertabular}
\usepackage{url}
\usepackage{hyperref}
\usepackage{threeparttable}

\title{\LARGE \bf
Rethinking Gaussian Trajectory Predictors: \\ Calibrated Uncertainty for Safe Planning
}

\author{Fatemeh Cheraghi Pouria, Mahsa Golchoubian, and  Katherine Driggs-Campbell
\thanks{Fatemeh Cheraghi Pouria and  Katherine Driggs-Campbell are with the Department of Electrical and Computer Engineering, University of Illinois at Urbana-Champaign, Champaign, IL 61820 USA (e-mail: fatemeh5@illinois.edu;
krdc@illinois.edu)
Mahsa Golchoubian is with the Department of Mathematical and Computational Sciences, University of Toronto, Toronto, Canada (e-email:mahsa.golchoubian@utoronto.ca)
\textit{(Corresponding author: Fatemeh Cheraghi Pouria.)}}
}

\usepackage{graphicx}
\begin{document}
\maketitle
\thispagestyle{empty}
\pagestyle{empty}

\begin{abstract}

\label{Abstract}Accurate trajectory prediction is critical for safe autonomous navigation in crowded environments. While many trajectory predictors output Gaussian distributions, the reliability of their confidence levels often remains unaddressed. This limitation can lead to unsafe or overly conservative motion planning when the predictor is integrated with an uncertainty-aware planner. Existing Gaussian trajectory predictors primarily rely on the Negative Log-Likelihood loss, which is prone to predict over- or under-confident distributions, and may compromise downstream planner safety. This paper introduces a novel loss function for calibrating prediction uncertainty which leverages Kernel Density Estimation to estimate the empirical distribution of confidence levels. The proposed formulation enforces consistency with the properties of a Gaussian assumption by explicitly matching the estimated empirical distribution to the Chi-squared distribution. To ensure accurate mean prediction, a Mean Squared Error term is also incorporated in the final loss formulation. Experimental results on real-world trajectory datasets show that our method significantly improves the reliability of confidence levels predicted by different State-Of-The-Art Gaussian trajectory predictors. We also demonstrate the importance of providing planners with reliable probabilistic insights (i.e., calibrated confidence levels) for collision-free navigation in complex scenarios. For this purpose, we integrate Gaussian trajectory predictors trained with our loss function with an uncertainty-aware Model Predictive Controller on scenarios extracted from real-world datasets, achieving improved planning performance through calibrated confidence levels.https://github.com/fate-79/Rethinking-Gaussian-Trajectory-Predictors.git

\end{abstract}

\section{INTRODUCTION}

Trajectory prediction is a core challenge for interactive autonomous systems \cite{rudenko2020human}. The importance of trajectory prediction is evident in the context of safe autonomous crowd navigation. When endowed with informative trajectory predictors, downstream planners can effectively maneuver through pedestrian crowds while avoiding collisions. 

Previously, researchers focused on deterministic models, which predict a single most-likely future trajectory represented as a sequence of points. However, the need for more realistic motion forecasting drew attention to the probabilistic and multi-modal nature of pedestrian behavior. As a result, later works utilized generative and probabilistic formulations, such as bivariate Gaussian models and Gaussian Mixture Models (GMMs) which enable generation of multiple plausible future trajectories through sampling~\cite{alahi2016social,Mohamed2020SocialSTGCNNASA, chen2025dstigcn, lv2023skgacn, li2025pedestrian, chen2024imgcn, sighencea2023d, chen2024stigcn}. Such predictors facilitate uncertainty-aware planning by providing multiple future trajectories and associated probabilities.

As a common practice in the trajectory prediction literature, the assessment of whether a predictor outperforms others and consequently is favorable for integration with a planner is often based only on point-wise accuracy metrics such as Average Displacement Error (ADE) and Final Displacement Error (FDE), or their Best-of-N (BoN) variants for multi-modal probabilistic models~\cite{pellegrini2009you, le2024social, huang2025interaction}. However, a trajectory predictor that excels only under these metrics provides little insight into the overall quality and structure of the predicted distribution. Since the predicted distribution itself encodes uncertainty for the planner, neglecting its structure collapses probabilistic forecasting into single-trajectory selection. This undermines the reliability required for safe uncertainty-aware planning in real-world scenarios.

Recognizing these limitations, recent works \cite{mohamed2022social, golchoubian2024uncertainty} have highlighted the importance of ensuring the statistical validity of probabilistic trajectory predictors, particularly for bivariate Gaussian models, which are among the most widely used representations in trajectory prediction~\cite{alahi2016social, chen2025dstigcn, sighencea2023d, chen2024stigcn, Mohamed2020SocialSTGCNNASA, lv2023skgacn, li2025pedestrian, chen2024imgcn }. They suggest that rather than focusing solely on point-wise accuracy, when developing a bivariate Gaussian model, one should care about a fundamental property of a bivariate Gaussian distributions~\cite{mohamed2022social, golchoubian2024uncertainty}: the squared Mahalanobis distance between the predicted mean and the true future position at a given confidence level follows a Chi-squared distribution with two degrees of freedom~\cite{erten2020combination}.
Since the confidence levels of the predicted distribution inform the planner about regions that are likely to contain the true future position, ensuring the validity of these predictions when developing a trajectory predictor is vital. To the best of our knowledge most State-Of-The-Art (SOTA) Gaussian trajectory predictors lack an evaluation of the predicted confidence levels, which raises the question of whether the predicted uncertainty is reliable. 

In this paper, we propose a model-agnostic loss function for training bivariate Gaussian trajectory predictors. Our proposed loss directly encourages the predicted confidence levels to follow the Chi-squared distribution. 
Our approach leverages Kernel Density Estimation (KDE) to estimate the empirical distribution of confidence levels, which we then match to the Chi-squared distribution. In addition, we incorporate a Mean Squared Error (MSE) loss term to encourage the predicted Gaussian mean to align closely with the true position. 
To validate our loss function, we integrate the trajectory predictors trained with our loss function into an uncertainty-aware Model Predictive Controller (MPC) and demonstrate the benefits of providing the planner with more reliable confidence estimates.

In summary, our key contributions are threefold: (i) We propose a novel loss function that improves the calibration of confidence levels in bivariate Gaussian trajectory prediction models 
(ii) We demonstrate that our method improves confidence level prediction on real-world trajectory datasets (iii) We highlight the importance of reliable confidence level estimation in motion planning by integrating our calibrated predictor with an uncertainty-aware MPC, demonstrating its advantage in scenarios derived from real-world data.

\section{Related Work}
\label{RelatedWorks}
\subsection{Pedestrian Trajectory Prediction }
Pedestrian trajectory prediction has been an active research area for decades, evolving from hand-engineered motion models~\cite{van2008reciprocal,helbing1995social,pouriatopology} to learning-based approaches~\cite{alahi2016social,huang2021learning,huang2020long,girase2021loki,chen2025dstigcn,Mohamed2020SocialSTGCNNASA}. Learning-based methods aim to capture the inherent complexity of human motion, which arises from factors such as multi-modality~\cite{chen2025dstigcn,chen2024imgcn}, environmental constraints~\cite{bai2025sceneaware}, social interactions~\cite{alahi2016social,Mohamed2020SocialSTGCNNASA}, and agent intent~\cite{huang2020long,girase2021loki}.

Many recent trajectory predictors model the uncertainty in human motion by predicting a distribution over possible future trajectories~\cite{alahi2016social, chen2025dstigcn, lv2023skgacn, li2025pedestrian, chen2024imgcn, sighencea2023d, chen2024stigcn,akhtyamov2023social, Mohamed2020SocialSTGCNNASA, golchoubian2023polar, golchoubian2024uncertainty, Zhang2024SpatialTemporalSpectralLAA, gupta2018social, dendorfer2021mg}.
These approaches can broadly be categorized into two groups: non-parametric generative-based predictors that produce multiple plausible future trajectories \cite{gupta2018social, dendorfer2021mg} and parametric probabilistic-based predictors trained via Maximum Likelihood Estimation (MLE) that output bivariate Gaussian distributions as the future steps of the trajectory~\cite{alahi2016social, Mohamed2020SocialSTGCNNASA, golchoubian2023polar, golchoubian2024uncertainty, Zhang2024SpatialTemporalSpectralLAA, chen2025dstigcn, lv2023skgacn, li2025pedestrian, chen2024imgcn, sighencea2023d, chen2024stigcn}. A common MLE-based strategy minimizes the Negative Log-Likelihood (NLL) loss to align the predicted distribution with assumed bivariate Gaussian distribution~\cite{ chen2025dstigcn, Mohamed2020SocialSTGCNNASA, chen2024stigcn,lv2023skgacn, alahi2016social, li2025pedestrian, chen2024imgcn, sighencea2023d}. However, recent studies have shown that optimizing NLL alone can lead to poorly calibrated uncertainty estimates and unreliable confidence regions~\cite{golchoubian2024uncertainty, mohamed2022social}. To address this limitation, alternative training objectives have been proposed that directly target the whole predicted distribution~\cite{golchoubian2024uncertainty, mohamed2022social}. These objectives aim to improve the reliability of predicted confidence levels, which is essential for uncertainty-aware planning. However, their designed objectives primarily encourage the model to minimize the predicted squared Mahalanobis distance. Yet, according to the Chi-squared distribution, the squared Mahalanobis distance is expected to span a wide range of values beyond zero. Our work introduces a new training objective which directly match the distribution of predicted squared Mahalanobis distance with Chi-square distribution to enhance the reliability of confidence levels produced by Gaussian trajectory predictors.

\subsection{Incorporating Prediction Uncertainty into Planning}
Predicting human trajectories is a fundamental requirement for autonomous planning in environments shared with pedestrians. Evidently, a considerable portion of the trajectory prediction research focuses on predicting pedestrian motion in pedestrian-vehicle mixed environments~\cite{golchoubian2023pedestrian}. 
While many approaches only incorporate the predicted trajectories into the planning framework~\cite{li2020socially, le2024social, chen2023social, katyal2020intent}, an increasing number of studies also consider the uncertainty associated with these predictions~\cite{golchoubian2024uncertainty, dixit2023adaptive, lindemann2023safe}.

In uncertainty-aware planning, it is common to associate each predicted trajectory point with a spatial region that is expected to contain the true future position with a specified probability. For probabilistic trajectory predictors, such as bivariate Gaussian models, these regions correspond to confidence levels defined by the predicted mean and covariance. The effectiveness of this representation therefore depends on the quality of the model prediction as miscalibrated confidence levels can lead to misleading uncertainty regions and adversely affect downstream planning. Prior work, such as~\cite{golchoubian2024uncertainty, mohamed2022social}, has highlighted that degradation in the confidence levels of bivariate Gaussian predictors harm the planner as the predicted uncertainty regions collapse to near-point estimates and fail to capture the ground truth.

To overcome this issue, some recent works~\cite{huang2025interaction, lindemann2023safe, muthali2023multi, dixit2023adaptive} adopt a post-processing approach based on conformal prediction~\cite{shafer2008tutorial} to construct circular uncertainty regions around predicted trajectory points~\cite{tumu2024multi,cleaveland2024conformal} and have shown promising results when the uncertainty regions are integrated into planning frameworks. However, a key limitation of conformal prediction is its reliance on sufficiently large datasets that can be partitioned into separate training, calibration, and test sets, which is often impractical with real-world trajectory datasets.
More fundamentally, post-hoc calibration does not improve the quality of the predicted uncertainty regions. Instead, greater benefit can be achieved by employing trajectory predictors that inherently produce well-calibrated uncertainty regions. This observation motivates our work, which focuses on enforcing uncertainty reliability directly during training, enabling Gaussian trajectory predictors to provide trustworthy confidence regions without requiring post-processing.

\section{Method}
\label{Method}
\label{method}

\subsection{Problem Formulation}
Considering a crowd scene including $N$ pedestrians,
let $X_t =\{(x_t^i,y_t^i)\}_{i=1}^N$ denote the true positions of all pedestrians at time $t$. Given the true trajectories during the observation horizon, $X_{obs} = \{X_1, ..., X_{T_{obs}}\}$, our trajectory predictor predicts bivariate Gaussian distributions over the subsequent positions throughout the prediction horizon denoted by $\hat{X}_{pred}=\{\hat{X}_{T_{obs}+1},...,\hat{X}_{T_{pred}}\}$, where each $(\hat{x}_t^i,\hat{y}_t^i) \in \hat{X}_t$ for $T_{obs}+1 \leq t\leq T_{pred}$ is modeled by a bivariate Gaussian, $(\hat{x}_t^i,\hat{y}_t^i) \sim \mathcal{N}(\mu_t^i, \Sigma_t^i)$.

\subsection{Properties of bivariate Gaussians and Pitfalls of NLL}
A bivariate Gaussian distribution has two main properties as a solution to a trajectory prediction problem. First, mean is the sample with the highest probability; thus, we want the predicted mean to align closely with the true position of the pedestrian. Second, confidence levels of a bivariate Gaussian encompass a specified proportion of the probability mass. More formally, for a random vector $ X\sim\mathcal{N}(\mu, \Sigma)$, the squared Mahalanobis distance of $X$ from the mean $\mu$, measured with respect to the covariance structure $\Sigma$ is given by $d^{MD}(X,\mu,\Sigma)^2= (X-\mu)^T\Sigma^{-1}(X-\mu)$ which follows a Chi-squared distribution with two degrees of freedom, $\mathcal{X}^2_2$~\cite{erten2020combination}. Therefore, the confidence level defined by:
\begin{align}
    \mathbb{P}\{(X-\mu)^T\Sigma^{-1}(X-\mu)\leq \mathcal{X}^2_2(\alpha)\} = 1-\alpha
    \label{eq: chi2}
\end{align}

encloses $(1 - \alpha) \times 100\%$ of the probability mass. Here, $\chi^2_2(\alpha)$ denotes the value of Chi-squared with two degrees of freedom evaluated at a significance level equal to $\alpha$. By predicting  a bivariate Gaussian distribution in a trajectory prediction problem,
we claim that the ground truth is somewhere within the $i^{th}$ confidence level of the distribution with the probability of $1-\alpha_i$. However, this claim is only valid when our predicted bivariate Gaussian distribution is "ideal". Given the aforementioned properties of mean and covariance of a bivariate Gaussian, we call a predicted bivariate Gaussian ideal if the predicted mean and covariance satisfy Eq~\ref{eq: chi2}. Otherwise, our confidence levels become unreliable meaning that they can be over-confident (under-confident) and  smaller (larger) since they are capturing a lower (higher) proportion of ground truth samples than theoretically expected. 

Trajectory predictors which predict bivariate Gaussian distributions mostly rely on NLL as the loss function~\cite{chen2025dstigcn, alahi2016social, Mohamed2020SocialSTGCNNASA}. 
\begin{align}
    \label{Eq:NLL}
    \mathcal{L}_{NLL} = \sum_{t=T_{obs+1}}^{T_{pred}} &\sum _{i =1}^{N} \log\big(\sqrt{2\pi\det(\Sigma_t^i)}\big)  \\
    &+  \frac{1}{2}\big((X_t^i-\mu_t^i)^T {(\Sigma_t^i)}^{-1}(X_t^i-\mu_t^i)\big) \nonumber
\end{align}

However, since our goal is to predict an ideal bivariate Gaussian distribution, relying solely on NLL loss can lead to potential issues in practice. First, Eq~\ref{Eq:NLL} reveals a potential imbalance in optimization. In practice, the model can reduce the second term in the loss more easily by focusing on minimizing the error between the predicted mean and the true position, rather than learning a meaningful covariance matrix.
Moreover, the first component of the loss encourages small covariance determinants, while the second component is better minimized with larger determinants. As a result, the model may fail to maintain a proper balance and lean toward one extreme, producing unrealistically small or large confidence intervals that do not accurately reflect true probabilities. 
Second, the squared Mahalanobis distance appears explicitly in the second term of the NLL formulation (Eq.~\ref{Eq:NLL}), incentivizing the model to reduce it toward zero to minimize the overall loss. This, however, may distort its expected distribution, causing it to deviate from the Chi-squared distribution, which naturally spans a broad range of values beyond zero.
In order to address the potential limitations of NLL and predicting bivariate Gaussian distributions satisfying the two aforementioned properties, we propose a new loss function that forces the network to match the distribution of the squared Mahalanobis distances with the Chi-squared distribution while minimizing MSE. The MSE term explicitly penalizes errors in the predicted mean, whereas in both the NLL loss and the proposed calibration term, the discrepancy between the predicted mean and the ground truth is coupled with the predicted covariance, preventing direct supervision of mean. 

\subsection{Proposed Loss Function}
We implement our proposed loss function on open-source bivariate Gaussian predictors that originally use NLL loss. This choice enables a fair comparison, as the network architecture remains unchanged. It is important to note that, as our contribution lies in the design of the loss function rather than the network architecture, the proposed approach is model-agnostic and can be applied to various neural network architectures.
The main objective of our proposed loss function is to ensure that the predicted distribution exhibits the properties of a bivariate Gaussian distribution as defined in Eq~\ref{eq: chi2}. 
To enforce this, we define a loss function based on the difference between the empirical distribution of squared Mahalanobis distances (computed from ground truth samples) and the theoretical Chi-squared distribution. Since we only have a limited number of ground truth samples, we approximate the empirical density using kernel density estimation (KDE), which provides a differentiable alternative to standard histograms, as follows:
\begin{align}
    \hat{p}(c_j|\bar{K},b, D_t^1,...,D_t^{N_t}) = \frac{1}{N_tb}\sum_{i=1}^{N_t}\bar{K}\bigg(\frac{c_j-D_t^i}{b}\bigg)
\end{align}
In this equation, $\{D_t^i\}_{i=1}^{N_t}$ denotes the set of random variables representing the squared Mahalanobis distance between ground truth positions and predicted distributions of all pedestrians at time $t$ where $N_t$ is number of pedestrians.
$\{c_j\}_{j=1}^k$ are uniformly spaced bin centers over the interval $[0, \max_i\{D_t^i\}]$, $\bar{K}$ is a kernel function, and $b$ is the bandwidth parameter. 
For our density estimation, we employ the standard normal distribution as our Gaussian kernel, followed by a softmax normalization with temperature parameter $T$ to obtain a smooth and differentiable estimation.
Finally, we sum up the probability densities to get the Cumulative Density Function (CDF), $\hat{F}(c_j)$, and  define our loss (see Figure~\ref{fig:CDF}), $\mathcal{L}_{CDF}$ as the mean absolute error between the estimated CDF, $\hat{F}(c_j)$, and Chi-squared CDF, $F_{\mathcal{X}_2^2}(c_j)$:
\begin{align}
      \mathcal{L}_{CDF} = \frac{1}{k}\sum_{j=1}^k |\hat{F}(c_j)-F_{\mathcal{X}_2^2}(c_j)|
\end{align}

Matching the distribution of confidence levels only guarantees to satisfy the second mentioned property of an ideal bivariate Gaussian. It alone does not guarantee that the predicted mean aligns with the ground truth. Thus, we add mean absolute error term between predicted means and ground truth positions, where $\beta $ controls its contribution to the overall loss:
\begin{align}
      \mathcal{L} =  \beta \mathcal{L}_{CDF} +  \frac{1}{N_t}\sum_{i=1}^{N_t}|\mu_t^i- X_t^i|
      \label{eq:CDF Loss}
\end{align}
\subsection{Model Predictive Control (MPC) Planner}
We aim to demonstrate the importance of predicting ideal confidence levels by integrating our trajectory predictor with an uncertainty-aware planner. The goal of the planner is to output the best control action at each time step $t$, denoted by $u_t$, given the dynamic of the system and a set of constraints. 
We employ MPC as our planning framework since its flexibility in defining stage costs and constraints allows the planner to take advantage of the predicted distribution and avoid collision. The following optimization problem formulates an uncertainty-aware variant of MPC, which we use to highlight the importance of ideal confidence levels. To account for uncertainty in human position predictions,
we adopt the collision avoidance constraint from \cite{Du2011} to ensure the MPC respects a minimum collision probability threshold, $p_{col}$, alongside a minimum required distance between robot and pedestrians, $d_{safe}$.
\begin{align}
\label{eq:MPC}
\operatorname*{minimize}_{p_{1:H},\,u_{0:H-1}}   
\sum_{t=0}^{H-1} \bigg( u_t^\top Q_u u_t
+& \sum_{i=1}^{N_t} \frac{Q_{MD}}{d_{t,i}^{MD}(p_t,\mu_{t}^i,\Sigma_{t}^i)} \\
+Q_P\big(\frac{\|p_t-p_{goal}\|}{\|p_0-p_{goal}\|}\big)^2\bigg)& \nonumber
+ Q_H\big(\frac{\|p_H-p_{goal}\|}{\|p_0-p_{goal}\|}\big)^2 \\
\operatorname*{Subject} \operatorname*{to} \quad \quad \quad \quad \quad \quad \nonumber \\
p_{t+1} = f(p_t, u_t) \\
d_{t,i}^{MD}(p_t,\mu_{t}^i,\Sigma_{t}^i)^2 
   &\geq -2 \ln \!\left(
     \sqrt{\det(2\pi\Sigma_{t}^i)}\frac{p_{col}}{A_{col}} 
   \right), \nonumber \\
d_{t,i}^{ED}(p_t,\mu_t^i)^2 
   &\geq (r_{rob}+r_{ped}+d_{safe})^2 \nonumber
\end{align}

In Eq. \ref{eq:MPC}, $H$ is the planning horizon, $N_t$ is the number of pedestrians, and $p_t$ is the robot position at time step $t$. $Q_u$, $Q_{MD}$, $Q_P$ and $Q_H$ are weights for control effort, Mahalanobis distance cost, stage goal cost, and terminal cost, respectively. $f(p_t,u_t)$ denotes the robot dynamics, which is set to unicycle model. $d^{ED}_{t,i}$ is the Euclidean distance between pedestrian $i$ and robot at $t$. $r_{rob}$ and $r_{ped}$ are the robot and pedestrians' radii and $A_{col} = \pi(r_{rob} + r_{ped})^2$ is the area of the safety zone around the robot that pedestrians must avoid to prevent collisions. $p_{goal}$ is the robot's goal position.

\section{Experiments}
\label{Experiments}

We conduct two sets of experiments: one to evaluate the predictor trained with our proposed loss function, and another to assess how prediction accuracy and confidence calibration affect motion planning within an uncertainty-aware planner.
\label{experiment}
\subsection{Dataset and Scenarios}
We use two publicly available pedestrian trajectory datasets: ETH \cite{pellegrini2009you} and UCY \cite{lerner2007crowds}. ETH dataset includes two scenes: ETH and HOTEL, while UCY dataset is composed of UNIV and ZARA scenes. 
We use 70\% of each dataset for training and the remaining for testing to ensure that training and evaluation are performed under the same data distribution. This in-distribution split allows us to train and assess calibration under matched statistical conditions, reducing the influence of distribution shift. In addition, to investigate transferability of our approach to out-of-distribution data, we also adopt the common leave-one-out strategy to train the newest baseline~\cite{chen2025dstigcn}. The Leave-one-out strategy involves reserving one dataset for testing while using the remaining datasets for training.
All scenes share a rate of 2.5 fps. Following all the baselines, each trajectory is observed for $3.2s$ ($T_{obs} = 8$) and the next $4.8s$ ($T_{pred} = 12$) are predicted. 

Motion planning scenarios are extracted from the test sets by introducing a robot into pedestrian crowds and assigning it a start and goal position. For real-world datasets, we extract 16-second scenarios (40 frames) with an 8-second interval between consecutive scenarios. For each scenario, a rectangular region enclosing the ground-truth trajectories of pedestrians is defined, within which four start-goal pairs are generated:
from the middle of the bottom/right edge to the middle of the top/left edge, and from the bottom-left/right corner to the top-right/left corner. 
The robot is modeled as a unicycle, and each scenario runs until the goal is reached or the robot freezes. In total, we generate 44, 64, 52, and 108 scenarios for ETH, HOTEL, UNIV, and ZARA, respectively.
\subsection{Baselines and Evaluation Metrics}
\textbf{Trajectory Prediction:} 
We compare five Gaussian trajectory predictors. In all experiments, we preserve the original model architectures and only replace the NLL loss with our proposed loss, ensuring a fair and controlled comparison:
1) Vanilla LSTM (VLSTM): an LSTM-based trajectory prediction architecture; 2) Social-LSTM (SLSTM): an LSTM-based architecture augmented with a social pooling layer~\cite{alahi2016social}; 3) Social-STGCNN: a spatio-temporal graph convolutional network that explicitly models social interactions~\cite{Mohamed2020SocialSTGCNNASA}; 4) DSTIGCN: a deformable spatio-temporal interaction graph network~\cite{chen2025dstigcn}; 5) Social-LSTM+MHD (SLSTM+MHD): the same architecture as SLSTM, trained by minimizing the NLL loss while additionally penalizing the Mahalanobis distance between the ground-truth positions and the predicted distribution~\cite{golchoubian2024uncertainty}.
Since SLSTM+MHD shares the exact architecture with SLSTM, we use the last baseline to conduct a three-way comparison among SLSTM, SLSTM+MHD, and $\text{SLSTM}^*$ (trained with our proposed loss) and study the effect of the loss function in isolation.

Model performance is evaluated by the following metrics:
\textit{Average Displacement Error (ADE)}: Average of Euclidean distance between predicted mean and ground truth position over the prediction horizon; 
\textit{Final Displacement Error (FDE)}: Average  Euclidean distance between the predicted mean and the ground truth position at the final time step of the prediction horizon; 
\textit{Delta Empirical Sigma Values (\textbf{$\Delta \mathbb{ESV}_i$})}: Difference between the empirical proportion of ground truth falling within the confidence level corresponding to $i^{th}$ standard deviation (e.g., $1\sigma$, $2\sigma$, $ 3\sigma$) of the predicted distribution and the corresponding proportion under an ideal bivariate Gaussian 
($\sigma_{ideal,1}=0.39$, $\sigma_{ideal,2}=0.86$ and $\sigma_{ideal,3}=0.99$). Negative and positive values indicate over-confident and under-confidence distribution respectively~\cite{ivanovic2022propagating};
\textit{Mean Absolute Delta Empirical Sigma Value (\textbf{$\overline{|\Delta \mathbb{ESV}|}$}}: Average of absolute Delta Empirical Sigma Values over 100 confidence levels. Values closer to zero are desirable.
\begin{figure}[!t]
\centering
\includegraphics[width=0.9\linewidth,height=0.18\textheight]{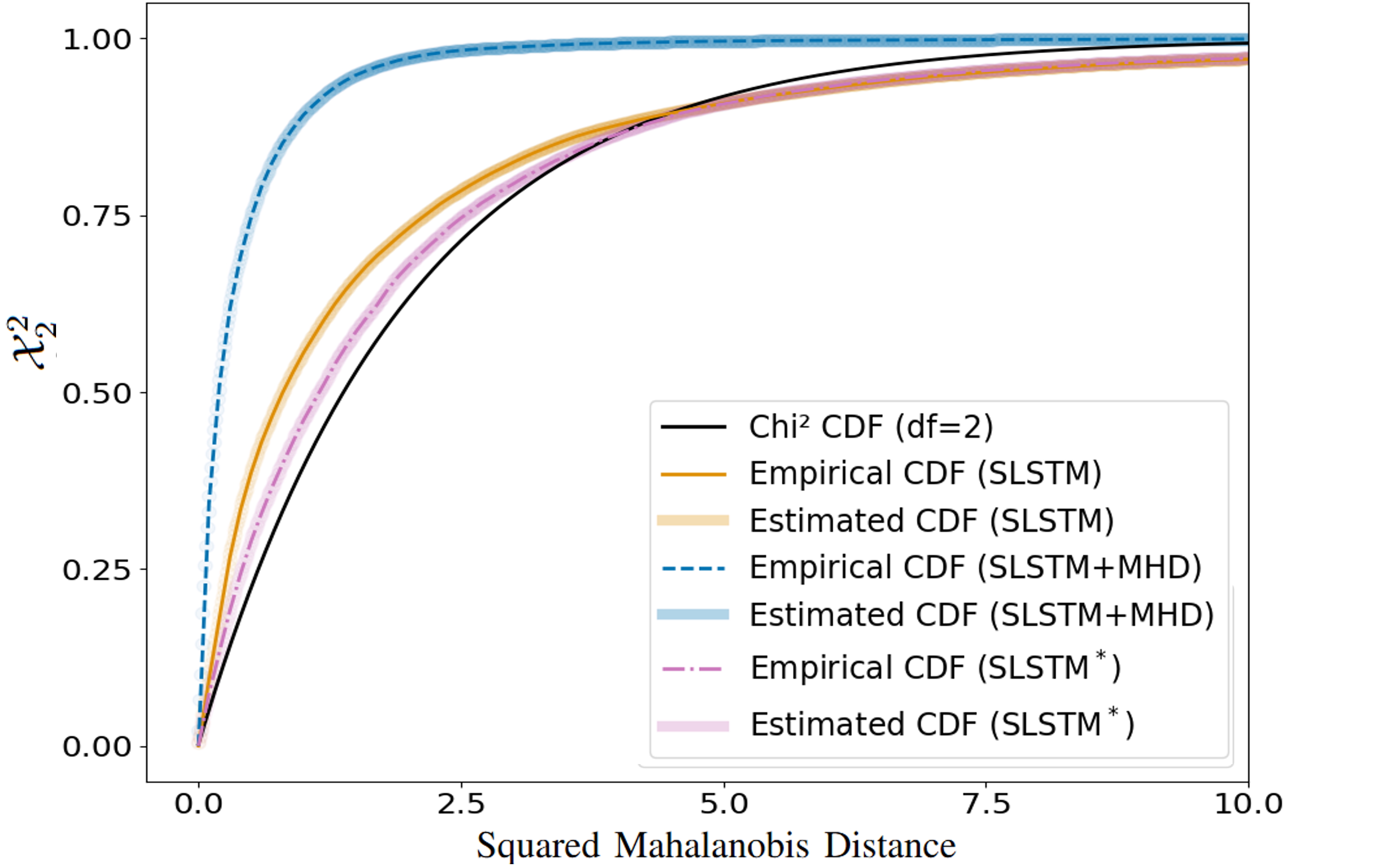}
\caption{Empirical (solid lines) and estimated (transparent lines) CDFs of the squared Mahalanobis distance predicted by three SLSTM variants on UNIV dataset.
The black curve denotes the theoretical $\chi^2_2$ distribution. Among the compared models, $\text{SLSTM}^*$ exhibits the closest alignment with the true distribution, indicating superior uncertainty calibration.} 
\label{fig:CDF}
\vspace{-10pt}
\end{figure}

\textbf{MPC Planner:}
The following metrics are used to evaluate the uncertainty-aware MPC planner integrated with different trajectory predictors. \textit{Success Rate (SR)}: percentage of the scenarios that the planner successfully reaches the goal; \textit{Collision Rate (CR)}: percentage of the scenarios that result in collision with pedestrians; \textit{Timeout Rate (TR)}: percentage of scenarios that planner freezes; \textit{Navigation Time (Nav. Time)}: number of steps that planner takes to achieve the goal. Each step is equal to 0.5 second; \textit{Path Length (Path Len.)}: length (in meters) of the path that planner takes to get to the goal; \textit{Percentage of Average Intrusion Ratio (Ave. Intr. Ratio \%):}  the percentage of time throughout the navigation in which the planner intrudes 
the personal space of any pedestrian where the personal space of each pedestrian is a circle centered at the pedestrian with radius of $0.6m$; \textit{Average Minimum Intrusion Distance (Ave. Min. Intr. Dist.)}: average of the closest distance between planner and pedestrian during intrusions.

\subsection{Implementation Details}
To preserve the model agnostic assumption of our method, we follow a pairwise comparison scheme where we keep the model architecture  of each baseline the same and train two variants: one on NLL loss and one on proposed loss. For the baseline model trained with the NLL loss, we retain the training hyperparameters reported in the original paper. In contrast, for the variant trained with our proposed loss function, we fine tune the training hyperparameters to achieve its best performance. We train the model minimizing our loss for 600 epochs using Adam optimizer \cite{kingma2014adam} with a learning rate of 0.001.
We set $\beta$ to 2 for VLSTM and Social-STGCNN and to 1 and 4 for SLSTM and DSTIGCN respectively. For each model, $\beta$ was fine-tuned to balance the relative magnitudes of the two loss terms.
For an accurate and high-resolution estimation of the squared Mahalanobis distances distribution, we use a step size of 0.01 between the centers of the histogram bins and we set both the temperature and bandwidth parameters in the kernel function to 0.005.
\vspace{-5pt}
\setlength{\tabcolsep}{5pt}
\begin{table}[h!]
\renewcommand{\arraystretch}{1.7}
\centering
\caption{MPC parameters}
\begin{tabular}{c|cccccccc}
Param& $Q_u$ & $Q_H$ & $Q_p$ & $Q_{MD}$& $r_{rob}$& $r_{ped}$ & $r_{safe}$ & $p_{col}$ \\
\hline
Value & $I_{2\times2}$ & $10^3$ & $10^3$ &$10^3$ &$0.2$ &$0.2$ & 0.4& 0.2 \\
\end{tabular}
\label{MPC_imp}
\end{table}
\vspace{-3pt}
Our MPC framework is built upon the work of \cite{akhtyamov2023social} where
the optimization problem is solved over a prediction horizon of $H = 12$, 
the control vector is bounded between $[0 m/s,-0.2 rad/s]$ and $[3 m/s, 0.2rad/s]$. The planner is initialized with a start velocity of $1m/s$ and angled towards the goal. 
The rest of MPC variables are listed in Table~\ref{MPC_imp}.

\section{Results}
\label{Results}
\subsection{Trajectory Prediction Results}

\begin{figure*}[tp]
\centering
\subfloat[]{\includegraphics[width=1.8 in]{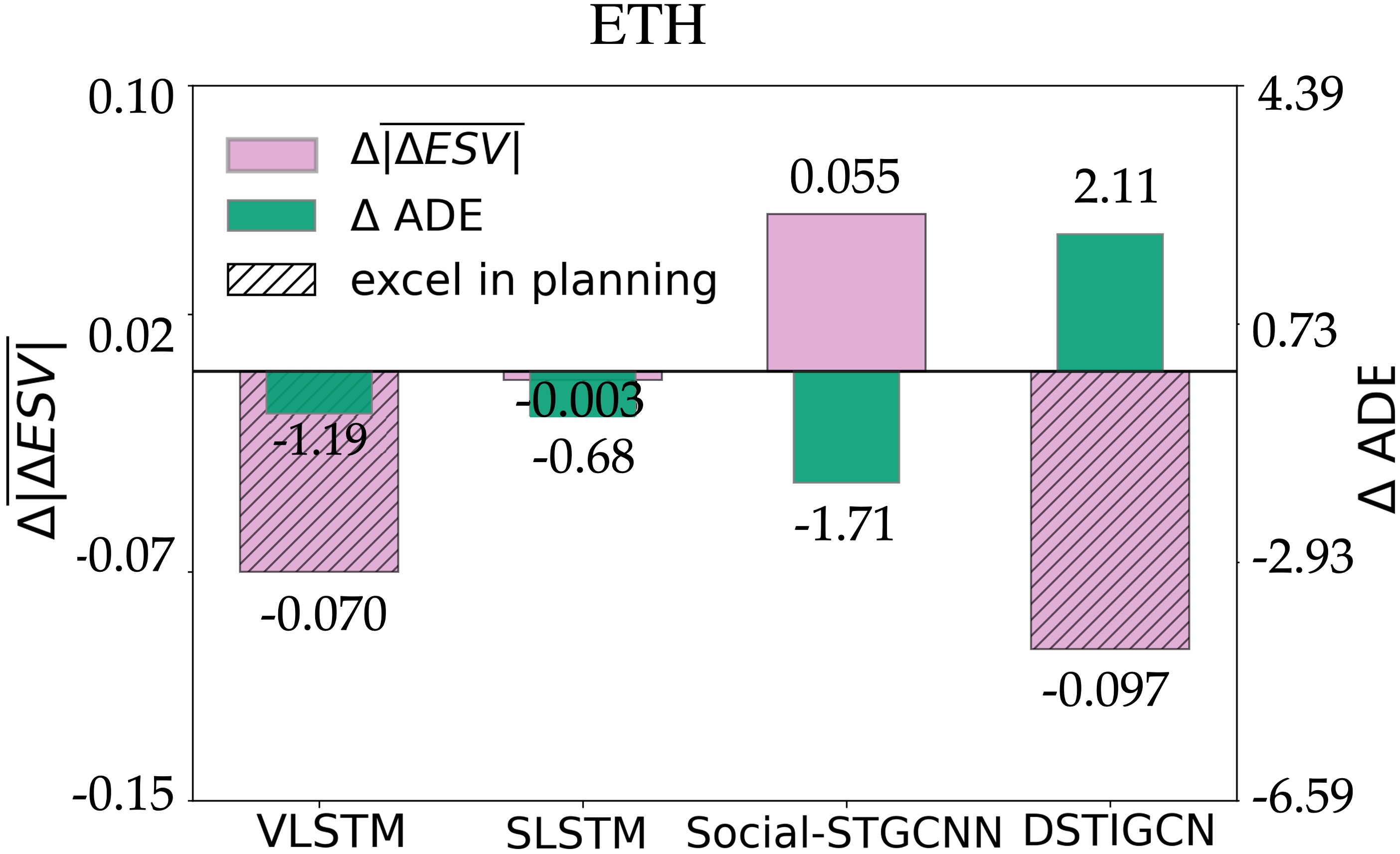}}%
\hfil
\hfil
\hfil
\subfloat[]{\includegraphics[width=1.63 in]{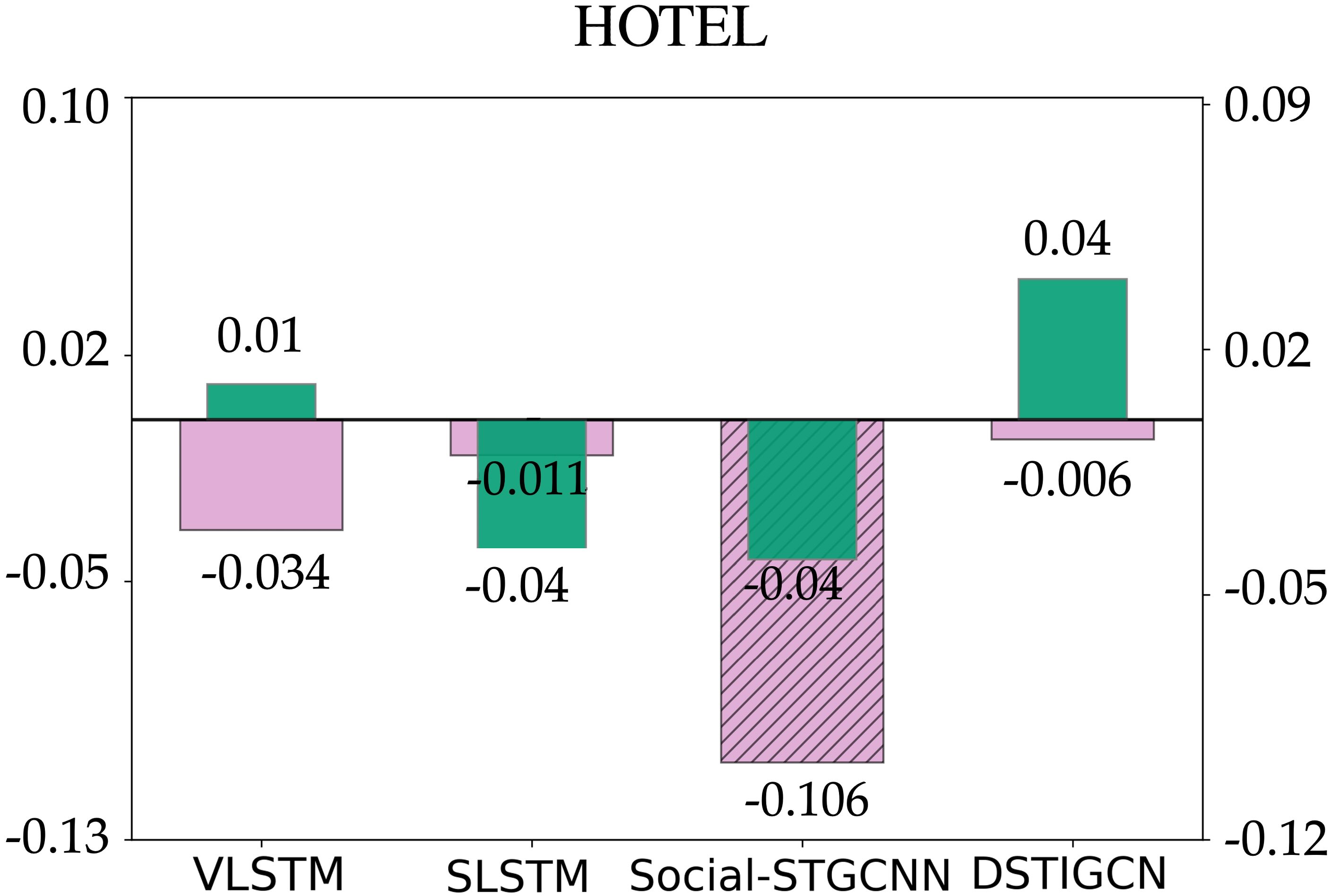}}%
\hfil
\hfil
\hfil
\subfloat[]{\includegraphics[width=1.67 in]{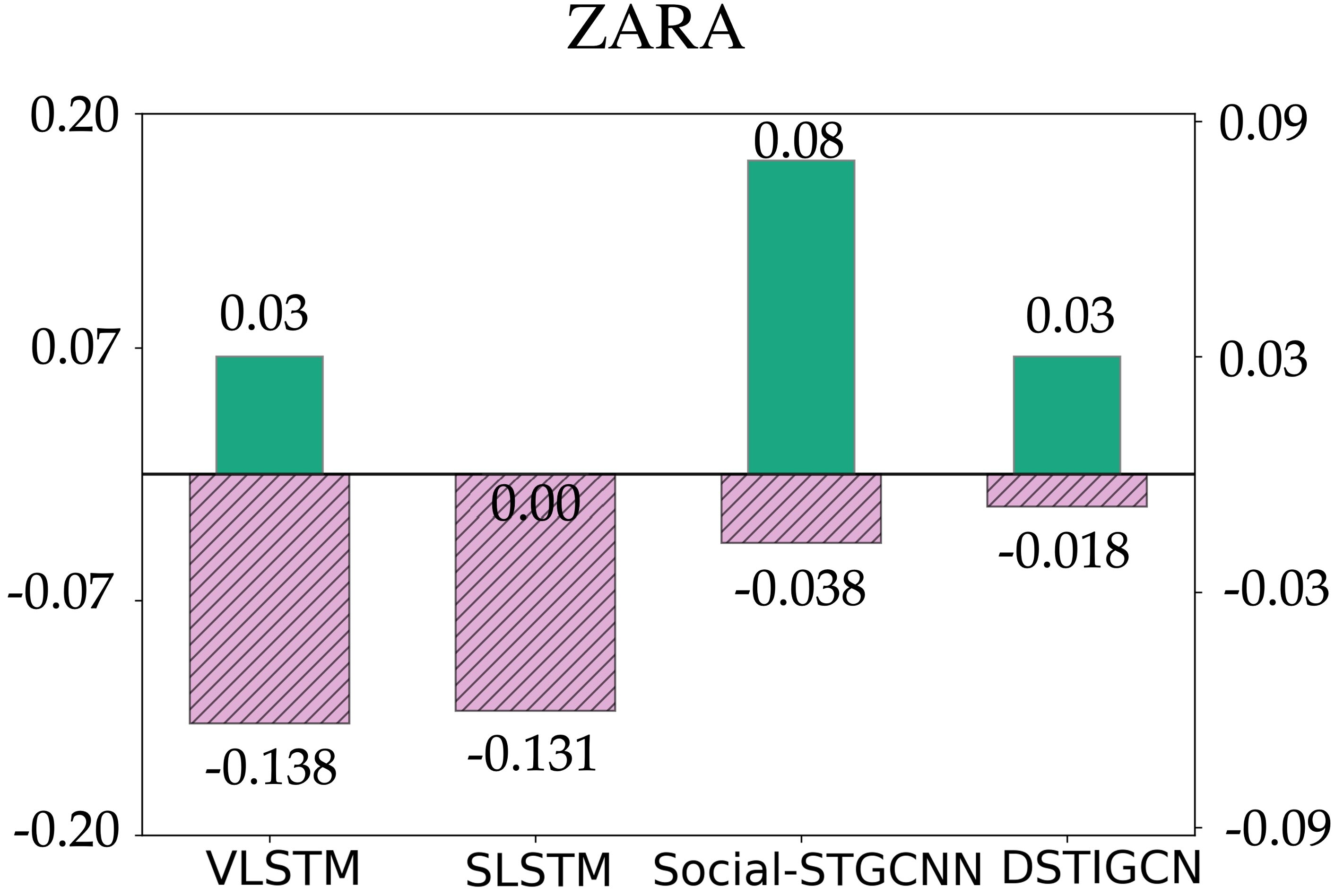}}%
\hfil
\hfil
\hfil
\subfloat[]{\includegraphics[width=1.67 in]{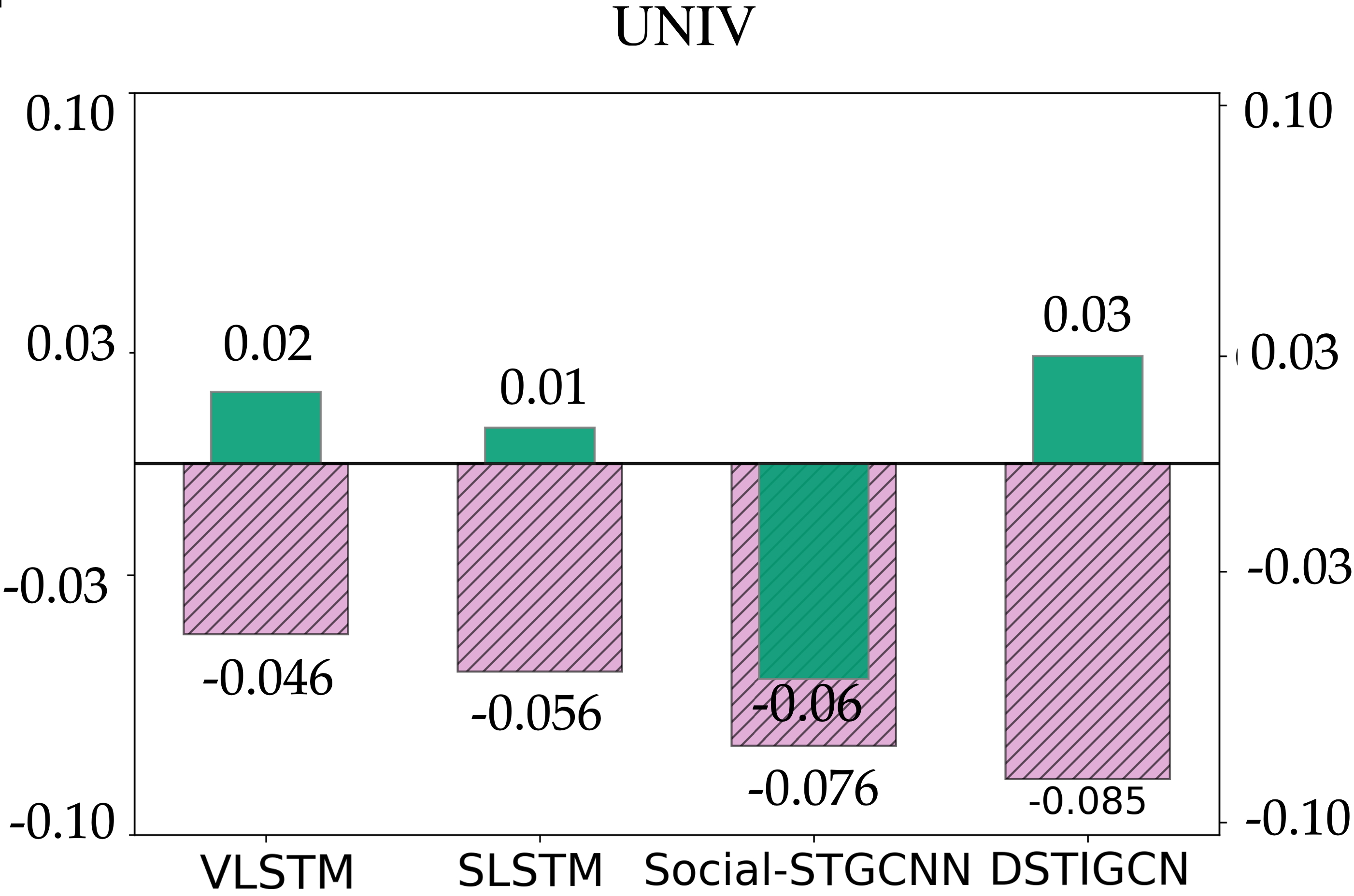}}
\caption{Difference in Mean Absolute Delta Empirical Sigma Value (\textbf{$ \overline{|\Delta \mathbb{ESV}|}$)} and Difference in Average Distance Error (ADE, unit:m) computed as: ``model trained with our loss - baseline model trained with NLL loss" for each model pair, on
(a) ETH, (b) HOTEL, (c) ZARA, (d) UNIV. More negative values indicate that the model trained with our loss obtains lower $\overline{|\Delta \mathbb{ESV}|}$ and ADE compared to the model trained with NLL. ($ \Delta \overline{|\Delta \mathbb{ESV}|}$) is hatched for cases where the trajectory predictor trained with our loss achieves a higher success rate in planning, indicating that gains in planning performance often coincide with significant improvement in $\overline{|\Delta \mathbb{ESV}|}$ resulting from our proposed loss function.}

\vspace{-5pt}
\label{fig: table}
\end{figure*} 

Table~\ref{table:tp MHD} reports trajectory prediction results for the SLSTM architecture trained with three different loss functions. 
The last four columns of Table~\ref{table:tp MHD} specifically assess the quality of the predicted confidence levels using $\Delta \mathbb{ESV}$ metrics. Averaged across all datasets, SLSTM architecture trained with our proposed loss achieves lower ADE/FDE as well as closer to zero $\Delta \mathbb{ESV}$ values, demonstrating the higher accuracy and well-calibrated confidence levels among compared approaches.

\setlength{\tabcolsep}{2.4pt}  
\begin{table}[htbp] 
\vspace{-8pt}
\begin{threeparttable}
\caption{Quantitative trajectory prediction results of three different variant of SLSTM trained with different loss}
\centering
\scriptsize
\begin{tabular}{llcccccc}
\toprule
\textbf{Dataset} & \textbf{Method} & $\textbf{ADE}$ / $\textbf{FDE}$& \textbf{$ \Delta \mathbb{ESV}_1$}  & \textbf{$\Delta \mathbb{ESV}_2$}   & \textbf{$\Delta \mathbb{ESV}_3$} & \textbf{$\overline{|\Delta \mathbb{ESV}|}$}\\
\midrule
\multirow{3}{*}{\textbf{ETH}} 
& SLSTM~\cite{alahi2016social}         & 2.26 / 3.74 &  \textbf{0.038} & -0.187 & -0.171 & 0.091\\
& SLSTM+MHD~\cite{golchoubian2024uncertainty}    & 3.51 / 5.23 & 0.070 & \textbf{-0.082} & -0.094 & \textbf{0.054}    \\
& $\text{SLSTM}^* $(ours)         &  \textbf{1.58} / \textbf{2.52} &  0.110 & -0.085 & \textbf{-0.089} & 0.088\\

\midrule
\multirow{3}{*}{\textbf{HOTEL}} 
& $\text{SLSTM}$         & 0.34 / 0.71  & \textbf{0.106} & -0.137 & -0.151 & 0.113 \\
& SLSTM+MHD  & 0.39 / 0.92 & 0.330 & 0.065 & -0.009 & 0.233 \\
& $\text{SLSTM}^*$         & \textbf{0.30} / \textbf{0.49} &  0.218 & \textbf{-0.045} & \textbf{-0.056} & \textbf{0.102} \\
\midrule
\multirow{3}{*}{\textbf{ ZARA}}
& $\text{SLSTM}$         & 0.42 / 0.78 & 0.280 & \textbf{0.021} & -0.044 & 0.177 \\
& SLSTM+MHD  & 0.42 / 0.81 & 0.512 & 0.109 & \textbf{-0.009} & 0.364 \\
& $\text{SLSTM}^*$         & \textbf{0.42} / \textbf{0.75} & \textbf{-0.045} & 0.023 & -0.030 & \textbf{0.046} \\
\midrule
\multirow{3}{*}{\textbf{UNIV}}
& $\text{SLSTM}$         &  \textbf{0.76} / 1.45 &  0.163 & 0.015 & -0.024 & 0.098 \\
& SLSTM+MHD & 0.77 / 1.46 &  0.500 & 0.133 & \textbf{0.009} & 0.342\\
& $\text{SLSTM}^*$         & 0.77 / \textbf{1.43} &  \textbf{0.067} & \textbf{0.001} & -0.022 & \textbf{0.042} \\
\midrule
\multirow{3}{*}{\textbf{AVE}}
& $\text{SLSTM}$         & 0.95 / 1.67 &  0.147 & -0.077 & -0.098 & 0.120  \\
& SLSTM+MHD & 1.27 / 2.11 & 0.353 & 0.056 & -0.026 & 0.248\\
& $\text{SLSTM}^*$         &  \textbf{0.77} / \textbf{1.30} &  \textbf{0.088} & \textbf{-0.027} & \textbf{-0.049} &  \textbf{0.070} \\
\midrule
\end{tabular}
\begin{tablenotes}
\footnotesize
\item[*]indicates the model trained with our proposed loss function.
\end{tablenotes}
\label{table:tp MHD}
\end{threeparttable}
\vspace{-5pt}
\end{table}

\setlength{\tabcolsep}{6pt}
\renewcommand{\arraystretch}{0.5}
\begin{table*}[htbp] 
\scriptsize
\centering
\begin{threeparttable}
\caption{Quantitative result of trajectory prediction and corresponding uncertainty-aware MPC planner.}
\begin{tabular}{ll cc|cccc}
\cmidrule(lr){3-8} 
& & 
\multicolumn{2}{c|}{\textbf{Trajectory Prediction}} & 
\multicolumn{4}{c}{\textbf{Uncertainty-aware Planning }} \\
\midrule
\textbf{Dataset (Scene Density)}& \textbf{Method} & $\textbf{ADE} $ / $\textbf{FDE} $ & \textbf{$\overline{|\Delta \mathbb{ESV}|}$} & \textbf{SR/CR/TR}  & \textbf{Nav. time}   & \textbf{Path. Len} & \textbf{$\%$Intr.Ratio }/\textbf{ Min.Intr.Dist.} \\
\midrule
\multirow{15}{*}{\textbf{ETH ($\textbf{0.16  ped/m}^2$)}}
& VLSTM         & 2.71 / 4.26  & 0.119 & 0.77 / 0.16 / 0.07 & \textbf{11.31$\pm$1.19} & \textbf{19.57$\pm$1.57} & 4.82$\pm$2.12 / 0.29$\pm$0.03  \\
& $\text{VLSTM}^*$(ours)  & \textbf{1.52} / \textbf{2.41} & \textbf{0.049} & \textbf{0.91} / \textbf{0.09} / \textbf{0.00} & 11.32$\pm$1.24 & 20.14$\pm$1.37 & \textbf{4.36$\pm$1.79} / \textbf{0.33$\pm$0.03}   \\
\cmidrule{2-8}
& SLSTM~\cite{alahi2016social}        &  2.26 / 3.74 &  0.091 & \textbf{0.86} / \textbf{0.09} / 0.05 & 12.78$\pm$1.24 & \textbf{20.20$\pm$1.43} & \textbf{4.35$\pm$1.65} / \textbf{0.38$\pm$0.02} \\
& $\text{SLSTM}^*$(ours)  & \textbf{1.58} / \textbf{2.52} & \textbf{0.088} & 0.82 / 0.16 / \textbf{0.02} & \textbf{11.81$\pm$1.25} & 20.86$\pm$1.60 & 5.33$\pm$1.95 / 0.31$\pm$0.04\\
\cmidrule{2-8}
& $\text{Social-STGCNN}$~\cite{Mohamed2020SocialSTGCNNASA}   & 4.08 / 6.80 & \textbf{0.107} & \textbf{0.95} / \textbf{0.05} / \textbf{0.00} & 13.49$\pm$1.31 & 22.78$\pm$1.59 & 2.93$\pm$1.99 / 0.36$\pm$0.04 \\
& $\text{Social-STGCNN}^*$(ours) & \textbf{2.37} / \textbf{4.22} & 0.162 & 0.93 / 0.07 / \textbf{0.00} & \textbf{11.55$\pm$1.08} & \textbf{19.88$\pm$1.22} & \textbf{2.17$\pm$0.67} / \textbf{0.39$\pm$0.04}\\
\cmidrule{2-8}
& $\text{DSTIGCN}$~\cite{chen2025dstigcn}   & \textbf{1.14} / \textbf{2.31} &  0.157 & 080 / 0.20 / \textbf{0.00} & \textbf{5.81$\pm$0.37} & \textbf{15.80$\pm$0.94} & 4.92$\pm$1.18 / \textbf{0.34$\pm$0.03}   \\
& $\text{DSTIGCN}^*$(ours) & 3.25 / 5.53 & \textbf{0.060} & \textbf{0.91} / \textbf{0.09} / \textbf{0.00} & 12.01$\pm$1.33 & 20.61$\pm$1.48 & \textbf{2.74$\pm$1.09} / 0.28$\pm$0.03  \\
\midrule
\multirow{15}{*}{\textbf{HOTEL ($\textbf{0.26  ped/m}^2$)}} 
& $\text{VLSTM}$         &  \textbf{0.36} / 0.76  & 0.106 & \textbf{0.91} / \textbf{0.09} / \textbf{0.00} & \textbf{4.24$\pm$0.29} & \textbf{9.90$\pm$0.57} & 8.04$\pm$1.55 / \textbf{0.41$\pm$0.02}\\
& $\text{VLSTM}^*$    &    0.37 / \textbf{0.56} &  \textbf{0.072} & 0.89 / 0.11 / \textbf{0.00} & 5.52$\pm$0.53 & 10.60$\pm$0.64 & \textbf{7.74$\pm$1.55} / 0.34$\pm$0.03  \\
\cmidrule{2-8}
& $\text{SLSTM}$         &  0.34 / 0.71 &  0.113 & \textbf{0.95} / \textbf{0.05} / \textbf{0.00} & \textbf{4.39$\pm$0.28} & \textbf{10.07$\pm$0.54} & 9.74$\pm$1.77 / \textbf{0.40$\pm$0.02} \\
& $\text{SLSTM}^*$    & \textbf{0.30} / \textbf{0.49} &  \textbf{0.102} & 0.91 / 0.09 / 0.00 & 5.08$\pm$0.40 & 10.19$\pm$0.57 & \textbf{7.99$\pm$1.59} / 0.35$\pm$0.02 \\
\cmidrule{2-8}
& $\text{Social-STGCNN}$   & 0.63 / 1.08 & 0.228 & 0.80 / 0.20 / \textbf{0.00} & \textbf{5.05$\pm$0.48} & \textbf{9.88$\pm$0.64} & 7.36$\pm$1.50 / \textbf{0.38$\pm$0.02}\\
& $\text{Social-STGCNN}^*$ &  \textbf{0.59} / \textbf{1.05} &  \textbf{0.122} & \textbf{0.86} / \textbf{0.14} / \textbf{0.00} & 5.63$\pm$0.49 & 10.01$\pm$0.62 & \textbf{5.80$\pm$1.29} / \textbf{0.38$\pm$0.03}\\
\cmidrule{2-8}
& $\text{DSTIGCN}$   & \textbf{0.23} / \textbf{0.44} &  0.073 & \textbf{0.94}  / \textbf{0.06} / \textbf{0.00} & \textbf{4.32}$\pm$0.34 & \textbf{9.91$\pm$0.54} & 10.63$\pm$1.82 / \textbf{0.42$\pm$0.01}   \\
& $\text{DSTIGCN}^*$ & 0.27 / 0.48  &  \textbf{0.067}  & 0.91 / 0.09 / \textbf{0.00}& 4.53$\pm$0.34 & 9.92$\pm$0.56 & \textbf{9.92$\pm$1.77} / \textbf{0.42$\pm$0.01}  \\
\midrule
\multirow{15}{*}{\textbf{ZARA ($\textbf{0.15  ped/m}^2$)}}
& $\text{VLSTM}$         & \textbf{0.41} / \textbf{0.77} & 0.185 & 0.80 / 0.18 / \textbf{0.02} & \textbf{9.26$\pm$0.81} & \textbf{13.29$\pm$0.75} & 20.18$\pm$2.39 / 0.39$\pm$0.00\\
& $\text{VLSTM}^*$         & 0.44 / 0.78 &  \textbf{0.047}  & \textbf{0.85} / \textbf{0.11} / 0.04 & 12.65$\pm$0.85 & 14.61$\pm$0.74 & \textbf{19.11$\pm$2.63} / \textbf{0.45$\pm$0.00}\\
\cmidrule{2-8}
& $\text{SLSTM}$         & \textbf{0.42} / 0.78 &  0.177 & 0.82 / 0.16 / \textbf{0.02} & \textbf{9.31$\pm$0.74} & \textbf{13.03$\pm$0.58} & \textbf{20.00$\pm$2.30} / \textbf{0.39$\pm$0.00}\\
& $\text{SLSTM}^*$         & \textbf{0.42} / \textbf{0.75} & \textbf{0.046}& \textbf{0.84} / \textbf{0.12} / 0.04 & 12.89$\pm$0.92 & 14.51$\pm$0.72 & 20.70$\pm$2.46 / \textbf{0.39$\pm$0.00} \\
\cmidrule{2-8}
& $\text{Social-STGCNN}$  &  \textbf{0.50} / \textbf{0.93} & 0.084 & 0.75 / 0.22 / \textbf{0.03} & \textbf{8.38$\pm$0.76} & \textbf{12.55$\pm$0.65} & 17.88$\pm$2.31 / 0.36$\pm$0.01\\
& $\text{Social-STGCNN}^*$ & 0.58 / 1.05 &  \textbf{0.046} & \textbf{0.77} / \textbf{0.20} / \textbf{0.03} & 8.42$\pm$0.71 & 13.03$\pm$0.62 & \textbf{12.81$\pm$1.82} / \textbf{0.41$\pm$0.01}\\
\cmidrule{2-8}
& $\text{DSTIGCN}$  & \textbf{0.32} / \textbf{0.67} &   0.089 & 0.81 / \textbf{0.15} / 0.04 & \textbf{7.82$\pm$0.65} & \textbf{12.01$\pm$0.56} & 23.26$\pm$2.29 / \textbf{0.42$\pm$0.01} \\
& $\text{DSTIGCN}^*$ & 0.35 / 0.69 & \textbf{0.071} & \textbf{0.84} / \textbf{0.15} / \textbf{0.01} & 10.08$\pm$0.82 & 13.67$\pm$0.64 & \textbf{19.97$\pm$2.26} / \textbf{0.42$\pm$0.00}\\
\midrule
\multirow{15}{*}{\textbf{UNIV ($\textbf{0.31  ped/m}^2$)}}
& $\text{VLSTM}$         &  \textbf{0.76} / \textbf{1.45}  & 0.081 & 0.83 / 0.17 / \textbf{0.00} & \textbf{18.47$\pm$0.88} & \textbf{30.47$\pm$1.37} & 8.71$\pm$1.76 / \textbf{0.39$\pm$0.01}  \\
& $\text{VLSTM}^*$         & 0.78 / 1.46 &  \textbf{0.035} & \textbf{0.90} / \textbf{0.10} / \textbf{0.00} & 20.21$\pm$0.81 & 31.26$\pm$1.55 & \textbf{4.78$\pm$1.58} / 0.38$\pm$0.02 \\
\cmidrule{2-8}
& $\text{SLSTM}$         & \textbf{0.76} / 1.45 & 0.098 & \textbf{0.85} / 0.15 / \textbf{0.00} & 19.47$\pm$0.91 & \textbf{31.03$\pm$1.51} & 5.84$\pm$1.41 / 0.38$\pm$0.00 \\
& $\text{SLSTM}^*$         &  0.77 / \textbf{1.43} & \textbf{0.042} &  \textbf{0.85} / \textbf{0.13} / 0.02 & \textbf{19.44$\pm$0.94} & 31.37$\pm$1.46 & \textbf{5.58$\pm$1.68} / \textbf{0.41$\pm$0.01}  \\
\cmidrule{2-8}
& $\text{Social-STGCNN}$  & 0.72 / 1.40 &  0.132 & 0.71 / 0.29 / \textbf{0.00} & \textbf{16.53$\pm$1.06} & \textbf{28.11$\pm$1.16} & 14.94$\pm$2.16 / 0.39$\pm$0.01  \\
& $\text{Social-STGCNN}^*$ &   \textbf{0.66} / \textbf{1.30}  & \textbf{0.056} & \textbf{0.85} / \textbf{0.15} / \textbf{0.00} & 16.76$\pm$0.91 & 28.50$\pm$1.11 & \textbf{8.18$\pm$1.56} / \textbf{0.40$\pm$0.01}\\
\cmidrule{2-8}
& $\text{DSTIGCN}$  & \textbf{0.53} / \textbf{1.15} & 0.111 & 0.75 / 0.23 / \textbf{0.02} & \textbf{15.74$\pm$1.15} & \textbf{25.11$\pm$0.92} & 21.17$\pm$2.46 / 0.39$\pm$0.01\\
& $\text{DSTIGCN}^*$ & 0.56 / 1.17 & \textbf{0.026} & \textbf{0.77} / \textbf{0.21} / \textbf{0.02} & 16.15$\pm$1.06 & 26.33$\pm$1.16 & \textbf{12.16$\pm$1.47} / \textbf{0.40$\pm$0.01}\\
\midrule
\label{table:tp result}
\end{tabular}
\begin{tablenotes}
\vspace{-5pt}
\footnotesize
\item[*]indicates the model trained with our proposed loss function.
\item[+] please refer to Appendix~\ref{Appendix} for comprehensive trajectory prediction result of each baselines.
\end{tablenotes}
\end{threeparttable}
\vspace{-15pt}
\end{table*}

\begin{figure}[!tb]
\centering
\subfloat[]{\includegraphics[width=1.6 in]{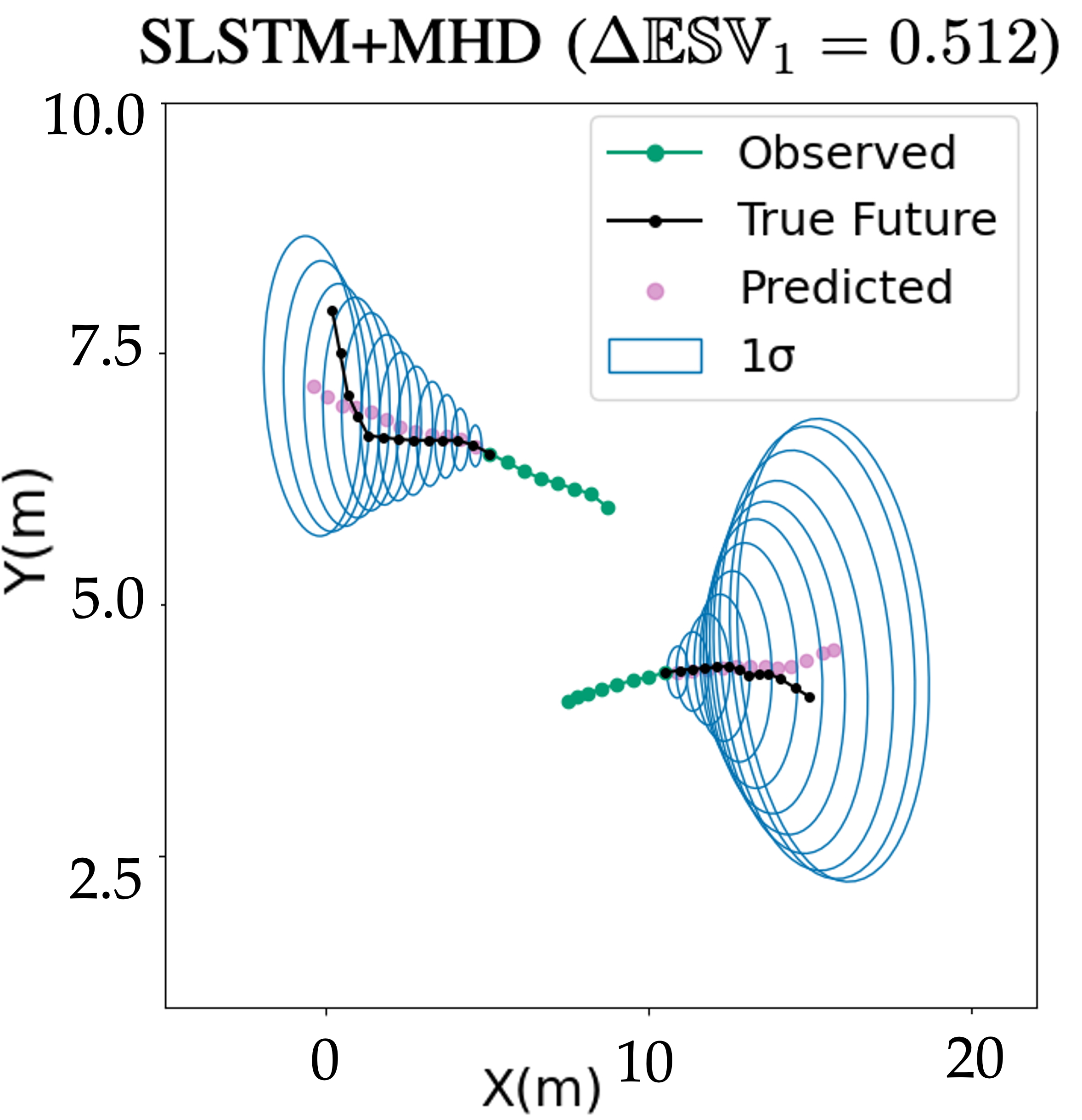}}%
\hfil
\hfil
\subfloat[]{\includegraphics[width=1.6 in]{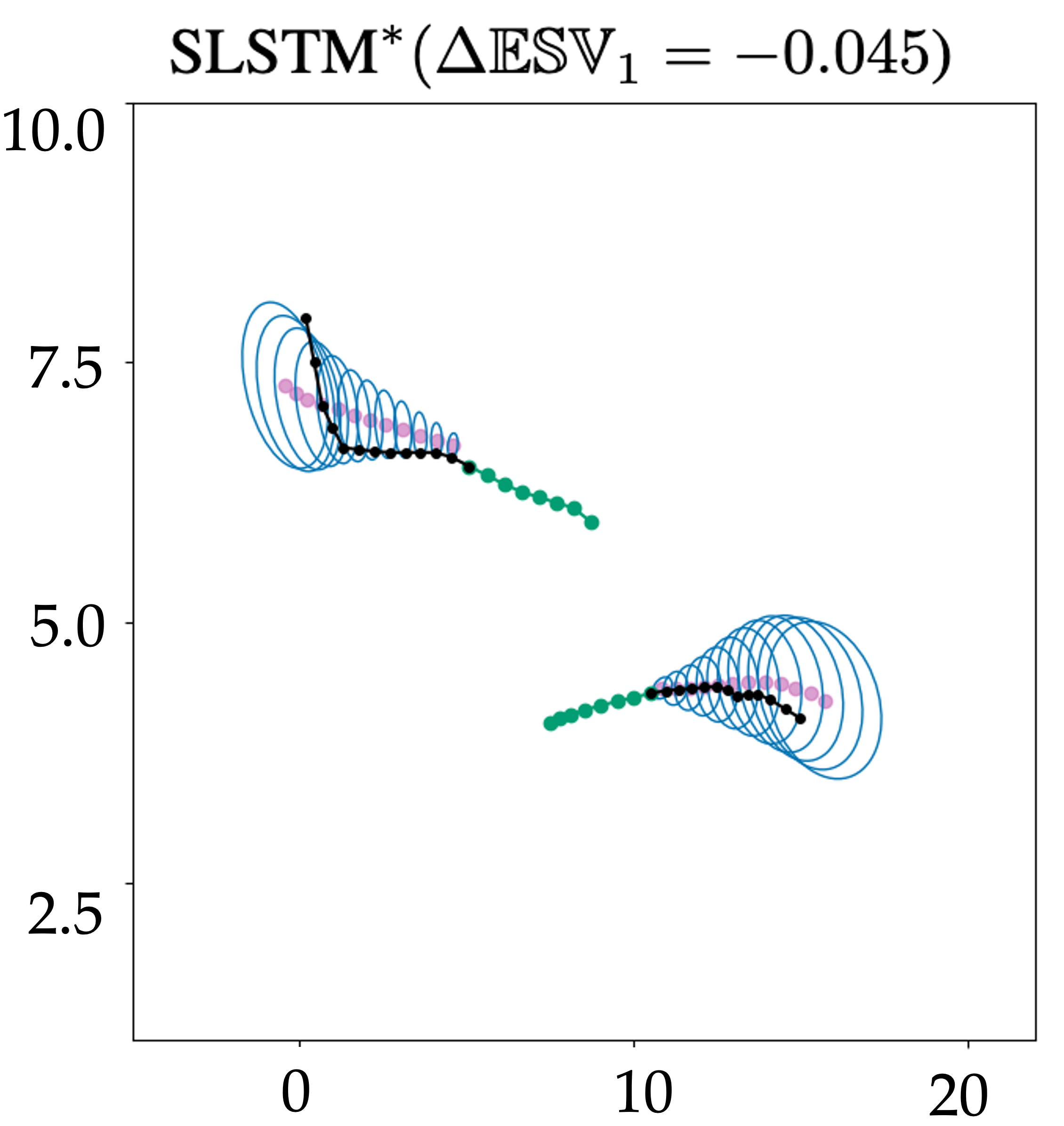}}%
\caption{Predicted trajectories and confidence levels corresponding to $1\sigma$ of (a) SLSTM+MHD and (b) $\text{SLSTM}^*$ evaluated on ZARA dataset. SLSTM+MHD has large positive \textbf{$ \Delta \mathbb{ESV}_1$}  which indicates under-confident behavior of the model and results in larger confidence levels than ideal; while \textbf{$\Delta \mathbb{ESV}_1$} of $\text{SLSTM}^*$ is the closer to the zero which makes its confidence levels to be the most similar to confidence levels of an ideal bivariate Gaussian.}
\label{fig: trajectory confidnce levels}
\vspace{-5pt}
\end{figure}

In contrast, SLSTM+MHD, which directly minimizes Mahalanobis distance, yields more positive $\Delta \mathbb{ESV}$ values, indicating under-confident predictions where the predicted confidence regions capture a larger proportion of ground truth than expected (Fig.~\ref{fig: trajectory confidnce levels}).
Meanwhile, SLSTM trained with the standard NLL shows more negative $\Delta \mathbb{ESV}_1$ and $\Delta \mathbb{ESV}_2$ values, corresponding to over-confident predictions.

Additionally, as shown in Fig.~\ref{fig:CDF}, we evaluate the fidelity of the predicted uncertainty by comparing the empirical distributions of squared Mahalanobis distances predicted by three SLSTM variants against the theoretical $\chi^2$ distribution. Evidently, the model trained with our proposed loss produces a distribution that closely follows the theoretical $\chi^2$ curve.  In contrast, the CDF of SLSTM+MHD deviates the most from the theoretical distribution and rises rapidly toward one at small squared Mahalanobis distances, indicating that many predictions concentrate near zero. This behavior is expected, as the loss function of SLSTM+MHD explicitly minimizes the Mahalanobis distance. The strong agreement between the empirical (solid lines) and estimated (transparent lines) distributions also supports the accuracy of our KDE method.

To show our approach is model-agnostic, we train several trajectory predictors with our proposed loss and compare them with their variant trained with the NLL loss. The results are summarized in Table~\ref{table:tp result} and Fig.~\ref{fig: table}.
Overall, our method reduces $\overline{|\Delta \mathbb{ESV}|}$ and lowers ADE/FDE in several cases, while any increases in ADE in other cases remain marginal.

We observe that the ADE/FDE on the ETH dataset are relatively large. This behavior stems from our experimental choice to train and test on the same dataset in order to remain within the same distribution and obtain more calibrated confidence levels. Due to the limited size of the ETH dataset, this setting leads to overfitting and prevents the models from sufficiently reducing ADE/FDE.

To further investigate our model performance on out-of-distribution datasets, we adopt a leave-one-out training strategy shown in Table~\ref{tp:OOD}, where the most recent model, DSTIGCN~\cite{chen2025dstigcn}, is trained with three datasets and evaluated on ETH. As shown in Table~\ref{tp:OOD}, the ADE degradation on ETH previously caused by overfitting due to the small dataset size is substantially mitigated. While the resulting $\Delta \mathbb{ESV}$ metrics are generally higher than those obtained under in-distribution training, our loss still yields improved $\Delta \mathbb{ESV}$ across most datasets and in average, highlighting the robustness of our approach under out-of-distribution evaluation.

\setlength{\tabcolsep}{3pt}
\begin{table}[!tb]
\centering
\begin{threeparttable}
\setlength{\tabcolsep}{4pt}
\caption{Quantitative trajectory prediction results of DSTIGCN under in-distribution and out-of-distribution training.}
\begin{tabular}{ll cc|cc}
\cmidrule(lr){3-6} 
& & 
\multicolumn{2}{c|}{$\textbf{In-Distribution}^+$} & 
\multicolumn{2}{c}{\textbf{Out-Of-Distribution}} \\
\midrule
\textbf{Dataset} & \textbf{Method} &
$\textbf{ADE} / \textbf{FDE} $ &
$\overline{|\Delta \mathbb{ESV}|}$ &
$\textbf{ADE} / \textbf{FDE} $ &
$\overline{|\Delta \mathbb{ESV}|}$ \\
\midrule
\multirow{2}{*}{\textbf{ETH}}
& DSTIGCN & \textbf{1.14} / \textbf{2.31} & 0.157 & \textbf{0.93} / \textbf{1.92} & 0.274 \\
& DSTIGCN$^{*}$ & 3.25 / 5.53 & \textbf{0.060} & 1.01 / 2.16 & \textbf{0.183} \\
\midrule
\multirow{2}{*}{\textbf{HOTEL}}
& DSTIGCN & \textbf{0.23} / \textbf{0.44} & 0.073 & \textbf{0.56} / 1.12 & 0.150\\
& DSTIGCN$^{*}$ & 0.27 / 0.48 & \textbf{0.067}  & \textbf{0.56} / \textbf{1.06} & \textbf{0.088} \\
\midrule
\multirow{2}{*}{\textbf{ZARA}}
& DSTIGCN & \textbf{0.32} / \textbf{0.67} & 0.089 & \textbf{0.36} / 0.79 & \textbf{0.103}  \\
& DSTIGCN$^{*}$ & 0.35 / 0.69 & \textbf{0.071} & \textbf{0.36} / \textbf{0.77} & 0.144\\
\midrule
\multirow{2}{*}{\textbf{UNIV}}
& DSTIGCN & \textbf{0.53} / \textbf{1.15} & 0.111 & \textbf{0.62} / \textbf{1.26} & 0.255 \\
& DSTIGCN$^{*}$ & 0.56 / 1.17 & \textbf{0.026} & 0.69 / 1.36 & \textbf{0.191}  \\
\midrule
\multirow{2}{*}{\textbf{AVE}}
& DSTIGCN & \textbf{0.56} / \textbf{1.14} & 0.109 & \textbf{0.62} / \textbf{1.27} & 0.196 \\
& DSTIGCN$^{*}$ & 1.11 / 2.25 & \textbf{0.056} & 0.66 / 1.34 & \textbf{0.152}\\
\bottomrule
\end{tabular}
\label{tp:OOD}
\begin{tablenotes}
\footnotesize
\item[*]indicates the model trained with our proposed loss function.
\item[+] borrowed from Table~\ref{table:tp result}
\end{tablenotes}
\end{threeparttable}
\vspace{-15pt}
\end{table}
\subsection{Effect of Calibrated Gaussian Predictions on Planning}

Under our uncertainty-aware MPC formulation (Eq.~\ref{eq:MPC}), uncertainty calibration (i.e., lower $\overline{|\Delta \mathbb{ESV}|}$) influences planning through the stage cost and safety constraints. Consequently, ADE alone is insufficient to evaluate a trajectory predictor, as lower ADE does not necessarily translate to better planning performance. We therefore evaluate each predictor in real-world planning scenarios to assess its downstream impact.

Table~\ref{table:tp result} reports the planning performance of each trajectory predictor on a fixed set of scenarios from the test portion of the real-world datasets.
We observe that higher planning success rate is achieved when confidence levels are better calibrated, meaning $\overline{|\Delta \mathbb{ESV}|}$ is sufficiently close to zero. In particular, models trained with our loss function that lead to a substantial reduction in $\overline{|\Delta \mathbb{ESV}|}$, attain higher success rates (see Fig.~\ref{fig: table}). In contrast, modest reductions in calibration error, such as for SLSTM on ETH and DSTIGCN on HOTEL, do not yield noticeable planning improvements

\begin{figure}[!tb]
\centering
\subfloat[]{\includegraphics[width=1.72 in]{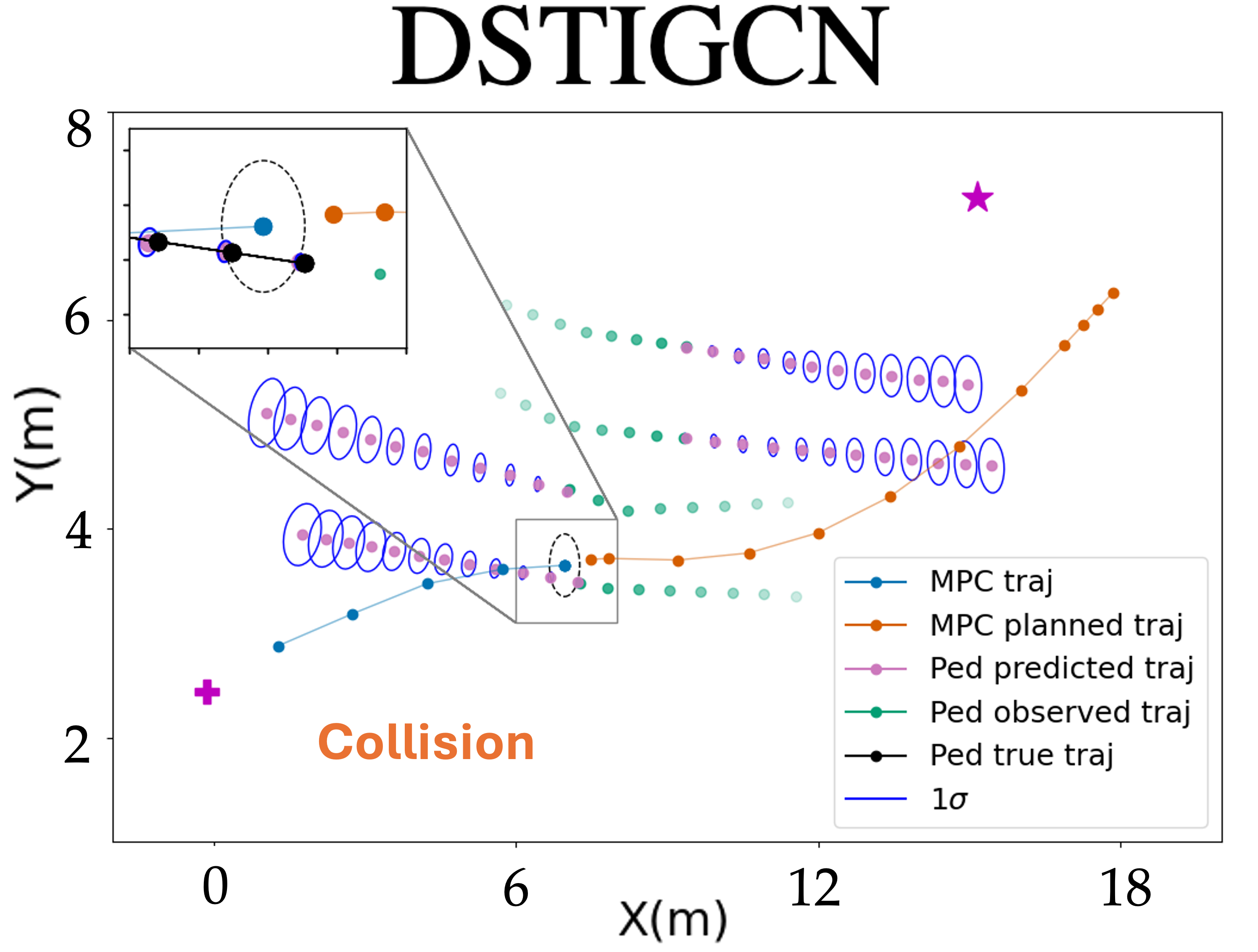}}%
\hfil
\hfil
\subfloat[]{\includegraphics[width=1.6 in]{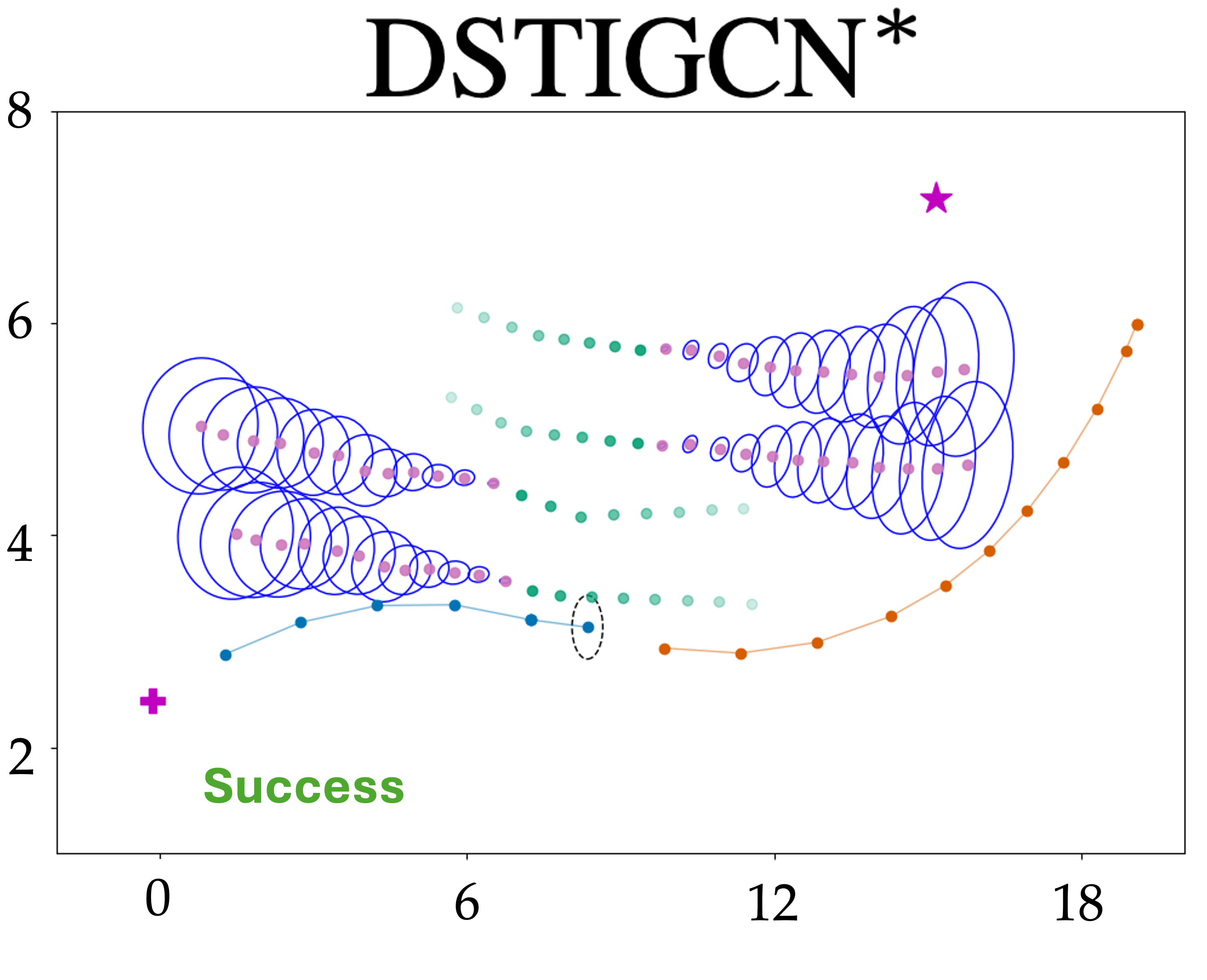}}
\caption{Planning result of DSTIGCN on a scenario from ZARA. Planner integrated with the model trained with our loss successfully reaches the goal by taking a longer collision-free path due to larger calibrated confidence level.} 
\label{planning}
\vspace{-17pt}
\end{figure}

A trade-off of using the trajectory predictors trained with our loss function is a moderate increase in navigation time and path length. Unlike models trained with the NLL loss, which are prone to produce over- or under-confident uncertainty estimates, models trained with our loss function predict better calibrated Gaussian distributions. As a result, compared to scenarios with mildly over-confident predictions, the planner may instead select longer detours or wait for pedestrians to pass in order to safely satisfy collision constraints (see Fig.~\ref{planning}). This safer behavior is beneficial, as it leads to a reduction in intrusion rate and promotes safer navigation.

\section{CONCLUSIONS}
\label{conclusion}

In this paper, we present a novel loss function for training bivariate  Gaussian trajectory predictors. This loss function is designed to address the limitations of NLL loss by producing reliable confidence levels through aligning the distribution of predicted squared Mahalanobis distances with the true Chi-squared distribution, consistent with Gaussian distribution properties. Through experiments we show the feasibility of integrating our proposed loss function with different SOTA predictors. The experimental results on real-world datasets demonstrate that bivariate Gaussian models trained with our loss function predict better calibrated confidence levels.
Moreover, we integrate two variants of each trajectory predictor, one trained with NLL loss and the other trained with our proposed loss, into an uncertainty-aware MPC planner and evaluate them on scenarios extracted from real-world datasets. Through these planning experiments, we demonstrate the importance of well-calibrated predicted uncertainty in achieving low-collision and reducing intrusion rates.
A possible direction for future
work is investigating and developing similar approach for other probabilistic predictors, such as GMMs, as all probabilistic models rely on distributional assumptions that may not perfectly match the underlying data. Since the proposed loss is model agnostic, it can also be applied to future trajectory prediction models, including emerging Transformer-based predictors.

\section*{ACKNOWLEDGMENT}
This work was supported by the National Science Foundation under Grant No. 2143435. We thank Dr. Zhe Huang for valuable discussions  during the early stages of this project and Dr. Neeloy Chakraborty for reviewing drafts of our work.



\setlength{\emergencystretch}{2em}
\bibliographystyle{IEEEtran} 
\vspace{-5pt}
\bibliography{references}

\onecolumn
\appendices
\section{Additional Results}
\label{Appendix}
We present comprehensive trajectory prediction results for each baseline. The tables below include \textbf{$ \Delta \mathbb{ESV}_1$}, \textbf{$\Delta \mathbb{ESV}_2$},  \textbf{$\Delta \mathbb{ESV}_3$} and $\overline{\textbf{ADE}} $ / $\overline{\textbf{FDE}}$  computed using the standard Best-of-$N$ (BoN) strategy with $N=20$, in addition to the metrics reported in Table~\ref{table:tp result}.

\begin{table*}[htbp] 
\caption{Quantitative result of trajectory prediction of VLSTM}
\centering
\scriptsize
\begin{tabular}{llcccccc}
\toprule
\textbf{Dataset} & \textbf{Method}  &  \textbf{ADE} $\downarrow$ / \textbf{FDE} $\downarrow$& $\overline{\textbf{ADE}} \downarrow$ / $\overline{\textbf{FDE}} \downarrow$& \textbf{$ \Delta \mathbb{ESV}_1$}  & \textbf{$\Delta \mathbb{ESV}_2$}   & \textbf{$\Delta \mathbb{ESV}_3$} & \textbf{$|\Delta \mathbb{ESV}|_{mean}$} $\downarrow$\\
\midrule
\multirow{2}{*}{\textbf{ETH}} 
& VLSTM         & 2.71 / 4.26 & 0.88 / 1.42 & -0.115 & -0.196 & -0.169 & 0.119 \\
& $\text{VLSTM}^* $  & \textbf{1.52} / \textbf{2.41} & \textbf{0.77} / \textbf{1.01} & \textbf{0.020} & \textbf{-0.112} & \textbf{-0.126} & \textbf{0.049}  \\
\midrule
\multirow{2}{*}{\textbf{HOTEL}} 
& $\text{VLSTM}$         &  \textbf{0.36} / 0.76 & \textbf{0.18} / \textbf{0.40} & \textbf{0.059} & -0.182 & -0.167 & 0.106  \\
& $\text{VLSTM}^* $    &    0.37 / \textbf{0.56} & 0.33 / 0.44 & 0.203 & \textbf{-0.017} & \textbf{-0.067} & \textbf{0.072}  \\
\midrule
\multirow{2}{*}{\textbf{ZARA}}
& $\text{VLSTM}$         & \textbf{0.41} / \textbf{0.77} & \textbf{0.19} / \textbf{0.36} & 0.266 & \textbf{0.013} & -0.041 & 0.185  \\
& $\text{VLSTM}^* $         & 0.44 / 0.78 & 0.23 / 0.41 & \textbf{-0.058} & 0.033 & \textbf{-0.019} & \textbf{0.047}  \\
\midrule
\multirow{2}{*}{\textbf{UNIV}}
& $\text{VLSTM}$         &  \textbf{0.76} / \textbf{1.45} & \textbf{0.31} / \textbf{0.60} & 0.134 & \textbf{-0.001} & -0.026 & 0.081  \\
& $\text{VLSTM}^* $         & 0.78 / 1.46 & 0.33 / 0.61 & \textbf{0.053} & 0.013 & \textbf{-0.014} & \textbf{0.035}  \\
\midrule
\multirow{2}{*}{\textbf{AVE}}
& $\text{VLSTM}$         &  1.06 / 1.81 & \textbf{0.39} / 0.70 & 0.086 & -0.092 & -0.101 & 0.123\\
& $\text{VLSTM}^* $         & \textbf{0.78} / \textbf{1.30} & 0.42 / \textbf{0.62} & \textbf{0.055} & \textbf{-0.021} & \textbf{-0.057} & \textbf{0.051} \\
\midrule
\end{tabular}
\vspace{-10pt}
\end{table*}

\begin{table*}[htbp] 
\caption{Quantitative result of trajectory prediction of SLSTM}
\centering
\scriptsize
\begin{tabular}{llcccccc}
\toprule
\textbf{Dataset} & \textbf{Method}  &  \textbf{ADE} $\downarrow$ / \textbf{FDE} $\downarrow$& $\overline{\textbf{ADE}} \downarrow$ / $\overline{\textbf{FDE}} \downarrow$& \textbf{$ \Delta \mathbb{ESV}_1$}  & \textbf{$\Delta \mathbb{ESV}_2$}   & \textbf{$\Delta \mathbb{ESV}_3$} & \textbf{$|\Delta \mathbb{ESV}|_{mean}$} $\downarrow$\\
\midrule
\multirow{2}{*}{\textbf{ETH}} 
& SLSTM         & 2.26 / 3.74 & 0.89 / 1.38 & \textbf{0.038} & -0.187 & -0.171 & 0.091\\
& $\text{SLSTM}^*$         &  \textbf{1.58} / \textbf{2.52} & \textbf{0.86} / \textbf{1.07} & 0.110 & \textbf{-0.085} & \textbf{-0.089} & \textbf{0.088}\\
\midrule
\multirow{2}{*}{\textbf{HOTEL}} 
& $\text{SLSTM}$         & 0.34 / 0.71 & \textbf{0.16} / 0.30 & \textbf{0.106} & -0.137 & -0.151 & 0.113 \\
& $\text{SLSTM}^*$         & \textbf{0.30} / \textbf{0.49} & 0.19 / \textbf{0.29} & 0.218 & \textbf{-0.045} & \textbf{-0.056} & \textbf{0.102} \\
\midrule
\multirow{2}{*}{\textbf{ZARA}}
& $\text{SLSTM}$         & 0.42 / 0.78 & \textbf{0.20} / \textbf{0.37} & 0.280 & \textbf{0.021} & -0.044 & 0.177 \\
& $\text{SLSTM}^*$         & \textbf{0.42} / \textbf{0.75} & 0.22 / 0.41 & \textbf{-0.045} & 0.023 & \textbf{-0.030} & \textbf{0.046} \\
\midrule
\multirow{2}{*}{\textbf{UNIV}}
& $\text{SLSTM}$         &  \textbf{0.76} / 1.45 & \textbf{0.31} / \textbf{0.59} & 0.163 & 0.015 & -0.024 & 0.098 \\
& $\text{SLSTM}^*$         & 0.77 / \textbf{1.43} & 0.32 / 0.60 & \textbf{0.067} & \textbf{0.001} & \textbf{-0.022} & \textbf{0.042} \\
\midrule
\multirow{2}{*}{\textbf{AVE}}
& $\text{SLSTM}$         & 0.95 / 1.67 & \textbf{0.39} / 0.66 & 0.147 & -0.077 & -0.098 & 0.120  \\
& $\text{SLSTM}^*$         &  \textbf{0.77} / \textbf{1.30} & 0.40 / \textbf{0.59} & \textbf{0.088} & \textbf{-0.027} & \textbf{-0.049} &  \textbf{0.070} \\
\midrule
\end{tabular}
\vspace{-10pt}
\end{table*}

\begin{table*}[htbp] 
\caption{Quantitative result of trajectory prediction of Social-STGCNN}
\centering
\scriptsize
\begin{tabular}{llcccccc}
\toprule
\textbf{Dataset} & \textbf{Method}  &  \textbf{ADE} $\downarrow$ / \textbf{FDE} $\downarrow$& $\overline{\textbf{ADE}} \downarrow$ / $\overline{\textbf{FDE}} \downarrow$& \textbf{$ \Delta \mathbb{ESV}_1$}  & \textbf{$\Delta \mathbb{ESV}_2$}   & \textbf{$\Delta \mathbb{ESV}_3$} & \textbf{$|\Delta \mathbb{ESV}|_{mean}$} $\downarrow$\\
\midrule
\multirow{2}{*}{\textbf{ETH}} 
& $\text{Social-STGCNN}$   & 4.08 / 6.80 &  \textbf{1.64} / 2.59 & \textbf{-0.140} & -0.104 & \textbf{-0.023} & \textbf{0.107}  \\
& $\text{Social-STGCNN}^*$   & \textbf{2.37} / \textbf{4.22} & 1.73 / \textbf{2.49}  & 0.296 & \textbf{-0.002} & -0.104 & 0.162   \\
\midrule
\multirow{2}{*}{\textbf{HOTEL}} 
& $\text{Social-STGCNN}$  &  0.63 / 1.08 & \textbf{0.37} / \textbf{0.47} & 0.329 & \textbf{0.064} & \textbf{-0.015} & 0.228  \\
& $\text{Social-STGCNN}^*$   & \textbf{0.59} / \textbf{1.05} & 0.45 / 0.67 & \textbf{0.106} & -0.172 & -0.171 & \textbf{0.122}  \\
\midrule
\multirow{2}{*}{\textbf{ZARA}}
& $\text{Social-STGCNN}$ & \textbf{0.50} / \textbf{0.93} & \textbf{0.30} / 0.51 &\textbf{0.013} & -0.178 & -0.161  &  0.084 \\
& $\text{Social-STGCNN}^*$   & 0.58 / 1.05 & 0.34 / \textbf{0.45} & 0.025 & \textbf{-0.098} & \textbf{-0.092} & \textbf{0.046}   \\
\midrule
\multirow{2}{*}{\textbf{UNIV}}
& $\text{Social-STGCNN}$  & 0.72 / 1.40 &  0.38 / 0.70 & -0.075 & -0.279 & -0.226 & 0.132   \\
& $\text{Social-STGCNN}^*$   & \textbf{0.66} / \textbf{1.30} & \textbf{0.32} / \textbf{0.48} & \textbf{0.041} & \textbf{-0.101} & \textbf{-0.083} & \textbf{0.056}  \\
\midrule
\multirow{2}{*}{\textbf{AVE}}
& $\text{Social-STGCNN}$  & 1.48 / 2.55 & \textbf{0.67} / 1.07 & \textbf{0.032} &-0.140 & \textbf{-0.106} & 0.138    \\
& $\text{Social-STGCNN}^*$   & \textbf{1.05} / \textbf{1.91} & 0.71 / \textbf{1.02} & 0.117 & \textbf{0.093} & 0.113 & \textbf{ 0.097}  \\
\midrule
\end{tabular}
\vspace{-10pt}
\end{table*}

\begin{table*}[htbp] 
\caption{Quantitative result of trajectory prediction  of DSTIGCN}
\centering
\scriptsize
\begin{tabular}{llcccccc}
\toprule
\textbf{Dataset} & \textbf{Method}  &  \textbf{ADE} $\downarrow$ / \textbf{FDE} $\downarrow$& $\overline{\textbf{ADE}} \downarrow$ / $\overline{\textbf{FDE}} \downarrow$& \textbf{$ \Delta \mathbb{ESV}_1$}  & \textbf{$\Delta \mathbb{ESV}_2$}   & \textbf{$\Delta \mathbb{ESV}_3$} & \textbf{$|\Delta \mathbb{ESV}|_{mean}$} $\downarrow$\\
\midrule
\multirow{2}{*}{\textbf{ETH}} 
& $\text{DSTIGCN}$   & \textbf{1.14} / \textbf{2.31} & \textbf{0.88} / \textbf{1.67} & \textbf{-0.072} & -0.357 & -0.367 & 0.157  \\
& $\text{DSTIGCN}^*$ & 3.25 / 5.53  & 1.40 / 2.27 & \textbf{-0.072} & \textbf{-0.075} & \textbf{-0.038} & \textbf{0.060}\\
\midrule
\multirow{2}{*}{\textbf{HOTEL}} 
& $\text{DSTIGCN}$   & \textbf{0.23} / \textbf{0.44} & \textbf{0.13} / 0.22 & \textbf{-0.011} & -0.181 & -0.138 & 0.077   \\
& $\text{DSTIGCN}^* $ & 0.27 / 0.48  & 0.15 / \textbf{0.20} & 0.106 & \textbf{-0.012} & \textbf{-0.045} & \textbf{0.067}  \\
\midrule
\multirow{2}{*}{\textbf{ZARA}}
& $\text{DSTIGCN}$  & \textbf{0.32} / \textbf{0.67} & 0.20 / 0.40 & \textbf{0.025} & -0.189 & -0.183 & 0.089 \\
& $\text{DSTIGCN}^* $ & 0.35 / 0.69 &  \textbf{0.19} / \textbf{0.30} & 0.098 & \textbf{0.039} & \textbf{-0.022} & \textbf{0.071} \\
\midrule
\multirow{2}{*}{\textbf{UNIV}}
& $\text{DSTIGCN}$  & \textbf{0.53} / \textbf{1.15} & 0.28 / 0.56 & \textbf{-0.039} & -0.244 & -0.199 & 0.111 \\
& $\text{DSTIGCN}^* $ & 0.56 / 1.17 &  \textbf{0.24} / \textbf{0.39} & 0.042 & \textbf{0.001} & \textbf{-0.019} & \textbf{0.026} \\
\midrule
\multirow{2}{*}{\textbf{AVE}}
& $\text{DSTIGCN}$  & \textbf{0.56} / \textbf{1.14} & \textbf{0.36} / \textbf{0.71} & \textbf{-0.024} & -0.243 & -0.222 & 0.109\\
& $\text{DSTIGCN}^* $ &  1.11 / 2.25 & 0.50 / 0.79 & 0.044 & \textbf{-0.012} & \textbf{-0.031} & \textbf{0.056} \\
\midrule
\end{tabular}
\end{table*}


\end{document}